\documentclass[letterpaper, 10 pt, conference]{ieeeconf}  

\IEEEoverridecommandlockouts                              

\usepackage{graphicx}
\usepackage{float}
\usepackage{caption}
\usepackage{subcaption}
\usepackage{svg}
\usepackage{wrapfig}
\usepackage{xurl}
\usepackage{hyperref}
\usepackage{multicol}
\DeclareCaptionType{equ}[][]
\usepackage{amsmath}
\usepackage{amssymb}
\usepackage{eucal}
\usepackage{algorithm}
\usepackage{algpseudocode}
\let\labelindent\relax
\usepackage{enumitem}
\usepackage{tabularx}
\usepackage{booktabs}
\usepackage{multirow}
\usepackage{xcolor}
\usepackage{lipsum}

\usepackage{graphicx}

\usepackage{tikz}
\usepackage{textcomp}
\usepackage{collcell}
\usepackage{color, xcolor, colortbl}
\usepackage{arydshln}
\usepackage{enumerate}
\definecolor{Gray}{gray}{0.9}
\definecolor{DarkGray}{gray}{0.8}
\definecolor{DarkerGray}{gray}{0.7}

\newcommand*{\opacity}{40}
\definecolor{high}{HTML}{32D732}
\definecolor{mid}{HTML}{FFBF00}
\definecolor{low}{HTML}{FF0000} 
\newcommand*{\minval}{0.00}
\newcommand*{\midval}{0.50}
\newcommand*{\maxval}{1.00}
\newcommand{\gradient}[1]{
    \ifdim #1 pt > \midval pt
            \pgfmathparse{int(round(100*(#1/(\maxval-\midval))-(\midval*(100/(\maxval-\midval)))))}
            \xdef\tempa{\pgfmathresult}
            \cellcolor{high!\tempa!mid!\opacity} #1
        \else
            \pgfmathparse{int(round(100*(#1/(\midval-\minval))-(\minval*(100/(\midval-\minval)))))}
            \xdef\tempa{\pgfmathresult}
            \cellcolor{mid!\tempa!low!\opacity} #1
        \fi
        }

\definecolor{red}{rgb}{1.00,0.00,0.00}
\definecolor{lightred}{rgb}{1.00,0.3,0.3}
\definecolor{blue}{rgb}{0.00,0.00,1.00}
\definecolor{green}{rgb}{0.1,0.50,0.1}
\definecolor{yellow}{rgb}{0.5,0.5,0.0}
\definecolor{white}{rgb}{1,1,1}

\newcommand{\cblue}[1] {\textcolor{blue}{\textbf{#1}}}

\title{\LARGE \bf
Agile and Versatile Robot Locomotion via Kernel-based Residual Learning
}

 \author{Milo Carroll$^{1}$, Zhaocheng Liu$^{1}$, Mohammadreza Kasaei$^{1}$ and  Zhibin~Li$^{2}$ 
\thanks{$^{1}$ Milo Carroll, Zhaocheng Liu and Mohammadreza Kasaei are with the School of Informatics, University of Edinburgh, UK. Email: \mbox{\{S2173175, zc.liu, m.kasaei\}@ed.ac.uk}}%
\thanks{$^{2}$ Zhibin~Li is with the Department of Computer Science, University College London, UK. Email: alex.li@ucl.ac.uk}%
}

\begin{document}

\maketitle
\thispagestyle{empty}
\pagestyle{empty}

\begin{abstract}
This work developed a kernel-based residual learning framework for quadrupedal robotic locomotion. Initially, a kernel neural network is trained with data collected from an MPC controller. Alongside a frozen kernel network, a residual controller network is trained via reinforcement learning to acquire generalized locomotion skills and resilience against external perturbations. With this proposed framework, a robust quadrupedal locomotion controller is learned with high sample efficiency and controllability, providing omnidirectional locomotion at continuous velocities. Its versatility and robustness are validated on unseen terrains that the expert MPC controller fails to traverse. Furthermore, the learned kernel can produce a range of functional locomotion behaviors and can generalize to unseen gaits.

\end{abstract}

\section{INTRODUCTION}

The versatility of legged locomotion exceeds other forms, such as wheeled locomotion, which requires continuous ground support and cannot feasibly adapt to challenging terrains \cite{survey_legged_1, legged_survey}. While quadrupedal animals access the most remote locations by exploring terrains that are never seen before \cite{model-free1-challenging-terrain}, other forms of robots usually would fail to do so. 

Traditional optimisation-based controllers perform well in challenging terrains \cite{mpc-cheetah, mpc-example2, mpc-example3}. However, due to high computation demands, they are prone to external perturbations and large model errors \cite{learning_to_walk, model-free1-challenging-terrain}. Recently, Deep Reinforcement Learning (DRL) methods have resulted in many robust locomotion controllers that operate at much higher frequencies enabling higher resiliency against errors and perturbations. However, RL-based controllers usually require carefully designed reward functions and excessive training data to produce an efficient controller with natural gaits \cite{gait-symetry1, gait-symetry2}. Additionally, the disagreement between physics simulators and the real world, DRL controllers also face the sim-to-real gap when testing on a real robot \cite{sim-to-real}. 

 Many legged animals start walking shortly after birth~\cite{mamals-walk-at-birth} due to pre-developed neural circuits, which are refined rapidly to acquire expert skills. Inspired by this, Residual learning~(ResL) is introduced for training RL agents only to adapt a prior control behavior, quickly learning robust and natural legged locomotion \cite{residual_origin,residual_physics_origin, residual_cassie, residual-cassie-2, cpg-based}. ResL methods can be grouped by the approach of providing the control priors: library-based, controller-based, and learning-based methods, of which have emerged chronologically; We break these down in the following subsections.  

\subsection{Residual Learning (ResL) Methods}

\textit{Library-based references}. 
These methods use pre-defined trajectory loops, which are static and can be quired to provide references~\cite{residual_cassie, residual-cassie-2}. They have been shown to produce robust and versatile locomotion in a sample efficient manner requiring less than 10M timesteps to converge. A library consisting of a single loop is all that is needed~\cite{residual_cassie}. However, improved velocity control can be achieved with a gait library providing trajectories queried by the target velocity. Yet, this can only provide priors for discrete velocities. Thus continuous velocity control requires the agent to work against the prior rather than working with them.

\begin{figure*}[!t]
    \centering
        \includegraphics[width=0.9\linewidth]{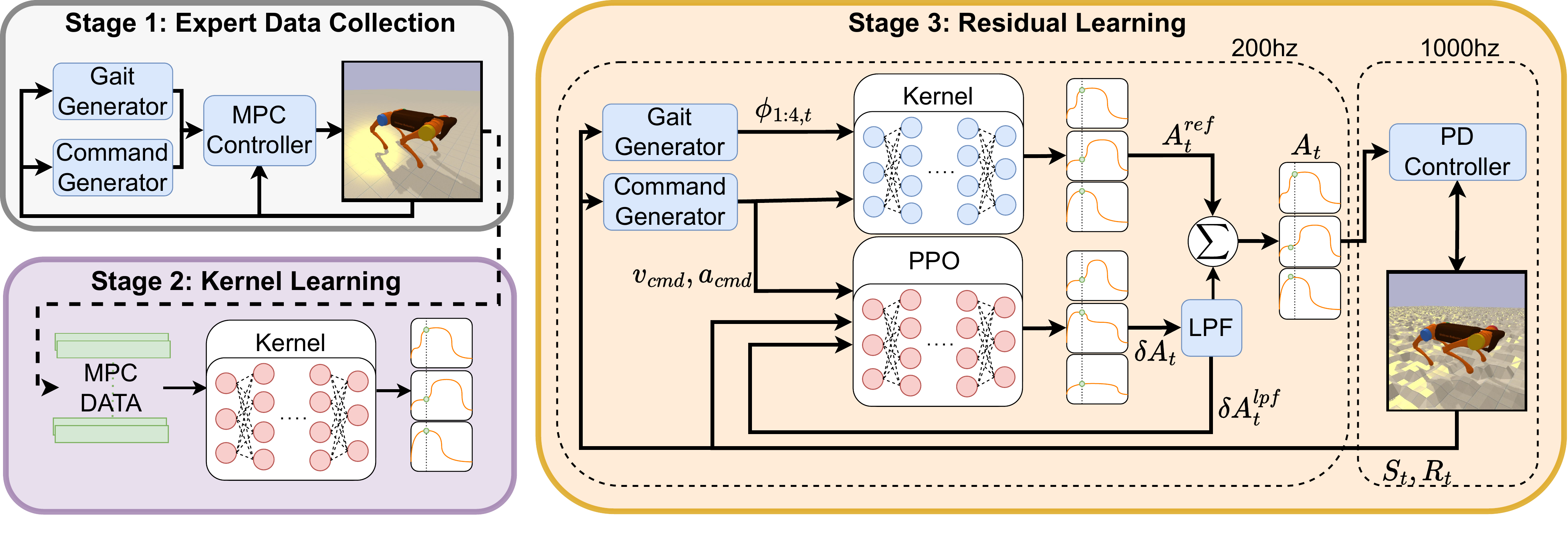}
        \vspace{-2mm}
    \caption{Overview of the proposed multi-stage robot locomotion framework, where the red components represent trainable modules, and blue components represent fixed modules.}
    \vspace{-5mm}
    \label{fig:arch:main}
\end{figure*}


\textit{Controller-based references}. These methods leverage existing expert controllers to provide the priors
within a ResL framework, \cite{cpg-based} and \cite{trajectory-adaption}, using MPC and CPG-based controllers, respectively. This is beneficial, as the expert controller provides omnidirectional locomotion priors with continuous velocity control. However, as the controllers are adaptive to the robots state, the residual agent must learn to model how these controllers respond, thus making the RL problem considerably more challenging. This is further reflected in the sample efficiency, with \cite{cpg-based} and \cite{trajectory-adaption} both requiring over 100M training timesteps to converge, considerably more than the library-based methods.

\textit{Learning references}. Learning has been incorporated into the reference generation process in a variety of ways \cite{cpg-resiual-linear-network, multi-modal-residual, vae_humanoid}. One approach uses a linear layer to adapt the trajectories produced by a CPG controller \cite{cpg-resiual-linear-network}, arguing this results in the production of more suitable trajectory priors for specific terrains. In~\cite{vae_humanoid}, a kernel is learned using a conditional variational auto-encoder (cVAE~\cite{cVAE}) from a motion database. The method provides the desired omnidirectional locomotion, velocity control, and versatility, even avoiding impossible terrain regions. Nevertheless, the sample efficiency of the method remains weak (200M). When priors are stochastic \cite{vae_humanoid} or adaptive \cite{cpg-based, trajectory-adaption}, we observe poor sample efficiency. We believe a gap is presented to produce sample efficiency similar to library-based methods, with the functionality comparable to controller-based methods by generating omnidirectional priors that provide continuous velocity control, are deterministic, and non-adaptive.

\subsection{Learning Trajectory-based Controllers}

Kinematic Motion Primitives (KMPs) \cite{kmps_theory, walk_trot_kmp} are used for developing data-driven locomotion controllers \cite{realized-kinematic-motion-primitives, kmp_humanoid}, but they produce static gaits and have no adaption, e.g.~walking at continuous target velocities \cite{kmp_humanoid}. Similar controllability problems exist in Dynamic Movement Primitives (DMPs) \cite{dmp_origin, dmp_survey, rhythemic_dmp_origin}. Although, a trained DMP's hyper-paramters can be tuned to adjust amplitude, frequency, and offset of the trajectories, showing potential for adaptive control. FastMimic \cite{FastMimic} exploits this, optimizing DMPs fitted to retargeted motion capture data, demonstrating rapid imitation learning on a physical robot \cite{source_mpc_controller_imiation_learning_rl}.

\par Discriminative Neural Networks (NN) are frequently used in trajectory prediction tasks but rarely within the locomotion domain.
In~\cite{autoencoder_quadraped_generalization}, an auto-encoder has been used to reconstruct the robot's state from a three-dimensional latent encoding; Given the reconstructed states, they can execute trajectory-based control. At inference time, \cite{autoencoder_quadraped_generalization} produce locomotion by injecting time dependant oscillatory latents ($\in [0,1]$) into the decoder, enabling the generation of unseen gait patterns but not locomotion. In~\cite{pretrained_rl}, a fully connected NN has been trained to predict trajectories given the robots' state. Despite achieving a low validation loss, functional locomotion is not observed due to the exclusion of time-dependent inputs. However, the model was trained to seed the NN of an RL agent, where functionality was not the primary concern. Generative models recently proposed have shown greater effectiveness. In~\cite{ox_vae_quadraped}, a cVAE has been used to develop a controller capable of navigating obstacles, gaps, and other challenging terrains. VAE-Loco \cite{vae-loco} uses a disentangled VAE \cite{dVAE} for trajectory prediction, producing an omnidirectional controller that controls the step height, frequency, and stance duration. However, as these methods are stochastic, they are not considered as a candidate solution.

In this paper, we approach the problem by providing deterministic, controllable, and learned priors, and thus bridge the gaps described in the aforementioned three ResL groups. Our core contribution is a novel ResL framework that is both sample efficient and highly controllable, providing omnidirectional locomotion at continuous velocities. Moreover, our framework is validated to be more robust and versatile than optimization-based controllers, and demonstrates considerably better performance in navigating across highly challenging terrains and robust responses to large perturbations.

The remained of this paper is organized as follows:  Section~\ref{sec:methodology} presents the proposed methodology. In Section~\ref{sec:simscenario}, a set of simulation environments for training and evaluating the framework will be designed. Following, Section~\ref{sec:results} conducts experiments to evaluate the performance of the proposed approach, discusses the findings, and compares the framework to other approaches. Finally, Section~\ref{sec:conclusion} concludes the core findings, weaknesses, and future research directions.

\section{Methodology}
\label{sec:methodology}

The locomotion framework detailed here enables omnidirectional locomotion, and demonstrates agile and versatile navigation across a broad range of unseen terrains. Given a target location, the controller must autonomously navigate a robot across  challenging terrains, such that the distance $D_{target}$ between the robot's position and is less than a minimum threshold $D_{min}$; maximizing the targets reached within a time limit. The following subsections presents a high-level view of our framework, and describes the non-parametric modules followed by a break-down our approach with control priors and the residual learning formulation.

\subsection{Overview of the Proposed Architecture}
The overall architecture of the proposed framework is depicted in Fig.~\ref{fig:arch:main}. As shown, it  contains a \textit{kernel}, a \textit{residual RL agent}, and a \textit{PD controller}. The kernel is an MLP trained to replicate the trajectories produced by a model-based MPC controller. Given a set of velocity commands, it outputs foot target positions in cartesian space relative to the robot's base. The RL agent learns to generate residual positional trajectories, learning the robot's dynamics and skills, such as balance recovery, providing agility and versatility to the framework. It produces foot target position deltas, summing with the kernel output to retrieve the final targets, as shown to be most effective by \cite{residual-cassie-2}.  The final foot target positions are converted into target joint angles using inverse-kinematics. The PD controller is responsible for generating torques applied to all joints that realize the desired behavior of the robot and attain the target joint angles.

\subsection{Analytical Components}

\textbf{Command Generator:}\label{sec:cmd_generator} We introduce a Command Gernerator module that generates X-Y and yaw velocity commands, given the robot's current location, $pos_{base}$, and orientation, $orn_{base}$, for chasing after a randomly sampled target location, $pos_{target}$. The velocity update at a frequency of $20$~hz, with a maximum delta of $\pm 0.005$. Velocity commands are constrained with in the range X:$\pm0.5$, Y: $\pm0.2$, Yaw: $\pm\pi/4$.

\textbf{Gait Generator:}\label{sec:gait_generator} The gait generator, inspired by \cite{source_mpc_controller_imiation_learning_rl} and \cite{fast_efficient_gait_transitions}, produces a contact schedule according the internal parameters: leg phases \mbox{$\phi_{1:4} \in (0,1]$}, initial phases \mbox{$\theta_{1:4} \in (0,1]$}, swing ratio $r_{swing} \in (0,1]$, and stance duration $\tau_{stance}$. $\phi_{1:4} \in (0,1]$ are updated at each time-step ($200$hz). Step cycles consist of two states: \emph{stance} ($\phi_i > r_{swing}$), when the feet are in contact with the ground, and \emph{swing} ($\phi_i \leq r_{swing}$) when not. Given the initial phases and the current time, we calculate the current phases: 
\begin{align}
\tau_{swing} &= \tau_{stance}/(1-r_{swing}) r_{swing}, \label{eq:swing_duration} \\
\tau_{step} &= \tau_{stance} + \tau_{swing}, \label{eq:cycle_duration} \\
\phi_i &= \theta_i + (\tau / \tau_{step}) \ mod \ 1 . \label{eq:current_phase}
 \end{align}

Different gaits are mainly defined by $\theta_{1:4}$, which determine the coordination between legs. When using the MPC controller \cite{source_mpc_controller_imiation_learning_rl}, $\tau_{stance}$ and $r_{swing}$ must be tuned to produce feasible gait patterns. Gait parameters are in Table \ref{tab:mpc:gaits_parameters}. 
\begin{table}[H]
\vspace{-1mm}
\centering
\captionof{table}{\label{tab:mpc:gaits_parameters} Gait generator parameters for different gaits.} 
\begin{tabular}{| c | c | c | c | c | c | c |}
    \hline
     \rowcolor{gray!30} Gaits & $\theta_1$ & $\theta_2$ & $\theta_3$ & $\theta_4$ & $\tau_{stance}$ & $r_{swing}$ \\ \hline
     walk & 0. & 0.5 & 0.75 & 0.25 & 0.3 & 0.25 \\ \hline
     trot & 0.9 & 0.4 & 0.4 & 0.9 & 0.3 & 0.4 \\ \hline
     bound & 0.4 & 0.4 & 0.9 & 0.9 & 0.1 & 0.3 \\ \hline
\end{tabular}
\vspace{-1mm}
\end{table}
 
\textbf{PD controller:} \label{sec:pd_controller} We apply the torque control loop at $1000$hz, as shown to be effective in the prior work of MELA \cite{MELA}. The $K_p$ and $K_d$ parameters are in Table~\ref{tab:pd_params}.

\begin{table}[!ht]
\vspace{-1mm}
\centering
\begin{footnotesize}
\caption{\label{tab:pd_params} Parameters of the PD controller.}
\begin{tabular}{ | c | c | c | c |  }
\hline
    \rowcolor{gray!30} Gains & abductor & hip  & knee \\ \hline
    $K_p$ & 100  & 100  & 100 \\\hline
    $K_d$ & 1  & 2  & 2 \\ \hline
\end{tabular}
\end{footnotesize}
\vspace{-1mm}
\end{table}

\textbf{Low Pass Filter:} As in~\cite{cpg-based}, only the residual outputs of the agent, $\delta\mathcal{A}_t$, are parsed by the LPF as the kernel trajectories are feasible and smooth:
\begin{equation}
\label{eq:lpf}
     \delta\mathcal{A}^{lpf}_t = \alpha \delta\mathcal{A}_t + (1-\alpha)  \delta\mathcal{A}^{lpf}_{t-1},
\end{equation}
\noindent
where $\alpha$ is the smoothing factor, $\delta\mathcal{A}^{lpf}_t$ is the residual after passing through the LPF. Here setting $\alpha$=0.1 can sufficiently remove most of the noise. Some noise is beneficial for policy exploration and improves responsiveness during highly noisy instances where over-smoothing introduces bias. 

\subsection{Kernel}


\textbf{Training Labels:} During swing states, the MPC controller~\cite{mpc_code}, \cite{source_mpc_controller_imiation_learning_rl,fast_efficient_gait_transitions} uses Raibert Heuristics \cite{raibert_swing}, which generates positional target trajectories $p^{ref}_{swing}$. We use these as labels for the swing legs. During stance states, we use the foot positions $p'_{stance}$ after applying the motor torques generated by the MPC stance controller as the labels. 

\textbf{Network Inputs} include the leg phase variables and velocity commands. We use the transformed normalized phase~$|\phi_i|$~(\ref{eq:norm_phase}), forcing the phase greater than one during swing states; This differentiates swing and stance states in input space while allowing the network to generalize to different gaits using an alternative $r_{swing}$.
\begin{equation}
\label{eq:norm_phase}
    |\phi_i| = 
    \begin{cases}
      1 + (\phi_i/r_{swing}), & \text{if}\ \phi_i <= r_{swing} \\
      (\phi_i - r_{swing}) / (1-r_{swing}), & \text{otherwise}
    \end{cases}.
\end{equation}

We denote the previously described as \mbox{\textit{kernel-base}}. As it accepts all the leg phases, it models the relative leg phases; As such, it cannot predict alternative gait patterns. \textit{kernel-ind} overcomes this modeling each leg individually, passing a single leg phase, velocity commands, and a one-hot encoding referring to the target leg. The final variant, \textit{kernel-ext}, builds on kernel-base, additionally accepting step height and ride height commands. Note, in the data collection process for \textit{kernel-ext}, we randomly select either step height $\in [0.05, 0.18]$  (default: 0.1) or the ride height $\in [0.18, 0.28]$ (default: 0.24) before walking to a new target location.

Using Optuna~\cite{optuna} to perform hyper-parameter tuning with Bayesian Optimization, we find the best results using the hyper-parameters in Table~\ref{tab:kernel:hparams}.

\begin{table}[H]
\vspace{-2mm}
\centering
\begin{footnotesize}
\captionof{table}{\label{tab:kernel:hparams} Kernel hyper-parameters (All variants).} 
\resizebox{0.9\linewidth}{!}{  
\begin{tabular}{ | c | c | c | c |}
\hline
    \rowcolor{gray!30} LR & Linear LR decay & Dropout & Batch-norm   \\ \hline
     $0.0024,$ & $0.7$ & $5e-6$ & False  \\ \hline
      \rowcolor{gray!30} Network & Activation & Loss & Batch size\\ \hline
     (256x4) & ReLU & L1 & 200\\\hline
\end{tabular}}
\end{footnotesize}
\vspace{-3mm}
\end{table}


\subsection{Residual Agent}
The residual RL agent, outputs positional residuals with a maximum magnitude of 5cm in each dimension for each leg. We also note that the kernel-base variant is applied for these experiments. Table~\ref{tab:ppo:hparams} shows the PPO hyper-parameters selected via a random search.

\begin{table}[H]
\vspace{-1mm}
\centering
\begin{footnotesize}
\captionof{table}{\label{tab:ppo:hparams} PPO hyper-parameters.} 
\resizebox{0.9\linewidth}{!}{  
\begin{tabular}{| c | c | c | c | c |}
\hline
    \rowcolor{gray!30} LR & LR exp decay & Entropy & Epochs & Rollout  \\ \hline
     $1e-3$ & $1e-7$ & $5e-6$ & 10 & 20000 \\ \hline
     \rowcolor{gray!30} Batch size & FE & Actor & Critic & \\ \hline
     4000 & (128x2) & (128x1) & (641) & \\\hline
\end{tabular}}
\end{footnotesize}
\vspace{-4mm}
\end{table}

\begin{figure*}[!t]
    \centering
    \includegraphics[width=0.95\textwidth]{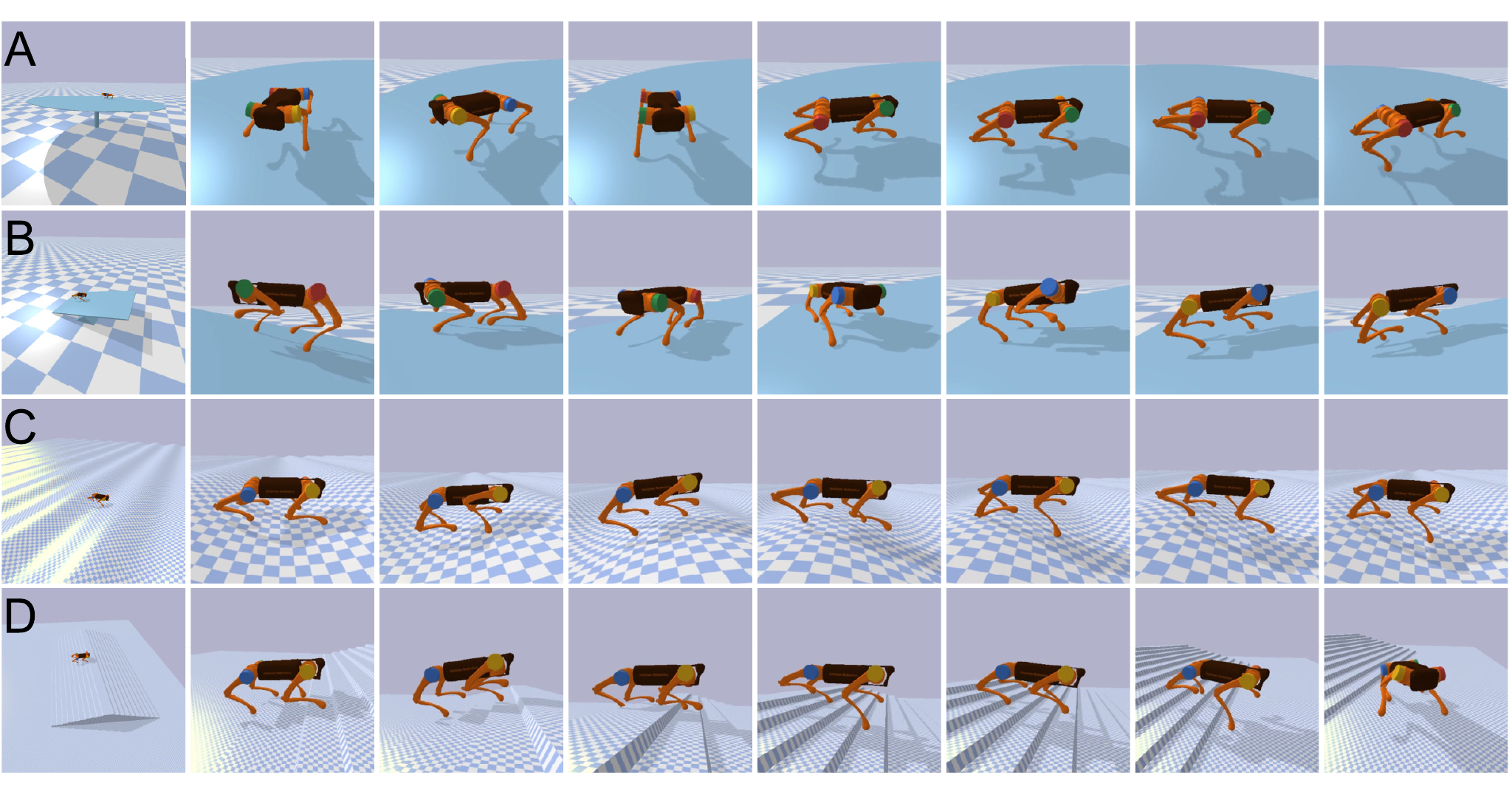}
    \caption{Evaluation of zero-shot task generalization on different terrains: (A) Tabletop: a 360-degrees see-saw platform with the maximum inclination angle of 5 degrees; (B) A seesaw table with maximum 6 degrees inclination angle; (C) sinusoidal surface; (D) Stairs on a flat ground.}
    \label{fig:eval_all_terrain}
    \vspace{-2mm}
\end{figure*}

\textbf{State space:} As opposed to other ResL methods providing deterministic priors \cite{residual_cassie, residual-cassie-2}, we find excluding the reference motion results in better learning. Although, we found improved performance passing the leg phases variables. Peak performance was achieved including neither, but passing the residual after passing through the LPF from the previous time-step $\delta A ^{lpf}_{t-1}$, which rectifies the Markov Property violation induced by using a LPF.

\begin{table}[H]
\vspace{-1mm}
\centering
\caption{\label{tab:rl:state_features_all} Residual RL agent state features.}
  \resizebox{0.98\linewidth}{!}{  

\begin{tabular}{| c | c | c |}
\hline
    \rowcolor{gray!30} State Feature & Description & Dimensions\\ \hline
    $v_{base}$ & Frontal, lateral, vertical velocities of the robots base & 3\\\hline
    $a_{base}$ & Roll, pitch, yaw velocities of the robots base & 3\\\hline
    $v_{cmd}$ & Target frontal and lateral velocities & 2\\\hline
    $a_{cmd}$ & Target yaw velocities & 1\\\hline
    $q$ & Joint angles & 12\\\hline
    $\dot{q}$ & Joint angles rotational velocities & 12\\\hline
    $CoM$ & Position of the center of the mass & 3\\\hline
    $pitch_{base}$ & Pitch of the robot base & 1\\\hline
    $roll_{base}$ & Roll of the robot base & 1\\\hline
    $fc_{1:4}$ & Contact state of each foot of the robot & 4\\\hline
    $\delta A^{lpf}_{t-1}$ & Residual applied at the previous time-step & 12\\\hline
\end{tabular}
 }
\end{table}

\textbf{Reward Function:} We use a mixture of radial basis functions~(RBF), $\phi_i(\gamma',\gamma,q) = \text{exp}(-(\gamma' - \gamma)^2  q)$, (shown to be effective in \cite{MELA, FastMimic}), and nominal rewards \mbox{$r_i \in \mathcal{F}_{nom}$} to define each feature of the reward function. RBF rewards function features, $\phi_i \in \Phi$, are parameterized by the target, $\gamma'$, and the curve steepness, $q$; A steeper RBF function incentivises learning and accommodates for attributes with small numeric errors. Equation~(\ref{eq:final_rewards}) represents our reward function and its parameters are summarized in Table~\ref{tab:rl:reward_function_final}.
\begin{equation}
\label{eq:final_rewards}
    R_t = \sum_{\phi_i \in \Phi} \omega_i  \phi_i(\gamma'_i, \gamma_i, q_i) + \sum_{r_i \in \mathcal{F}_{nom}} r_i .
\end{equation}

\begin{table}[t]
\vspace{-1mm}
\centering
\caption{Reward function parameters.\label{tab:rl:reward_function_final}}
\resizebox{0.95\linewidth}{!}{  
\begin{tabular}{| l | c | c | c | c |}
\hline
    \rowcolor{gray!30} $\phi_i \in \Phi$ & $\gamma'$ & $\gamma$ & $q$ & weight\\ \hline 

    Linear velocity  & $v_{cmd}$ & $v_{base}$ & 18.42 & 0.0076 \\  \hline
    Angular velocity & $a_{cmd}$ & $a_{base}$ & 7.47 & 0.0264\\ \hline
    Center of mass & $[0,0,-1]$ & $CoM$ & 2.35 & 0.0298 \\ \hline
    Distance to target & $0$ & $D_{target}$ & 0.74 & 0.0169\\ \hline
    Roll and Pitch & $[0,0]$ & $[pitch_{base}, roll_{base}]$ & 7.47  & 0.0298\\ \hline
    \rowcolor{gray!30} $r_i \in \mathcal{F}_{nom}$ & \multicolumn{3}{c|}{Reward function} & weight\\ \hline
    Falling penalty  &  \multicolumn{3}{c|}{
    $r = 
    \begin{cases}
      -19.8, & \text{if the robot fell} \\
      0, & \text{otherwise}
    \end{cases}$
    } & 1  \\ \hline
    Target reached   &  \multicolumn{3}{c|}{
    $r = 
    \begin{cases}
      8.75, & \text{if}\ D_{target} \leq D_{min} \\
      0, & \text{otherwise}
    \end{cases}$
    } & 1   \\ \hline
                         
\end{tabular}}
\vspace{-4mm}
\end{table}

\section{Simulations}
\label{sec:simscenario}

In this section, multiple scenarios for different aspect of training and evaluation of the kernel and residual agent will be designed. To this end, a simulated A1 uni-tree quadruped~\cite{a1} in the PyBullet~\cite{pybullet} physics simulator is used. In our simulations, we wrap the PyBullet simulation in an Open-Ai Gym~\cite{gym} environment during RL experiments.

\subsection{Training the framework}

The first stage of the training process, training the kernel, requires collecting locomotion data from an expert MPC controller. The MPC controller \cite{source_mpc_controller_imiation_learning_rl} executes the trot gait, with a stance duration of $0.2$~s, which reduces the variation in CoM allowing the network to learn better. It navigates to 500 consecutive target locations over the flat terrain, set at a minimum distance of 2.5m in a random direction. Collecting the data network inputs $\{ \ v_{cmd}, \ a_{cmd}, \ q, \ \phi_{1:4} \}$ before actions are taken, and labels $\{p^{ref}_{swing}, \ p'_{stance} \}$ and after each time-step ($200$hz).

In the second stage, we train the residual RL agent on randomly selected terrains for five consecutive episodes (75$\%$ height field, 25$\%$ perlin). The height-field perturbations are sampled uniformly $\sim \in [3\text{cm}, 4.5\text{cm}]$. Also, force perturbations are applied to the robot at a random point on the robot body, in a random direction horizontally, at intervals $\sim \in[5,8]$ seconds, with a magnitude $\sim\in[100,350]$N, for a duration of $0.3$s. The agent is trained for a total of 20M timesteps, tasked with navigating to randomly selected target locations, with a precision of $D_{min}=0.5m$, over 5 cpu's in parallel, taking roughly 8 hours (NVIDIA GeForce GTX 1060 6GB, AMD Ryzen 5 2600X Six-Core Processor).

\subsection{Evaluating the framework}

The framework is evaluated for it's versatility in four terrains in ascending difficulty: A)~Tabletop, B)~Seesaw, C)~Sinusoidal, and D)~Stairs  (see Fig.~\ref{fig:eval_all_terrain}). We set 5 target locations to reach per run, placed to challenge the agent, and start from 4 different starting locations. The pivoting tabletop has a maximum rotation around the pivot of 5deg. The seesaw has an decline/incline of 6deg. The stairs have a step height of 4cm. The sinusoidal terrain has a maximum incline of 11.5deg. 

Furthermore, we separately evaluate the robustness of the framework applying external forces to the robot. It is tasked with walking to a single target location on a flat terrain, where a force is applied to a random location on the robots body in a random direction in the horizontal plane for a duration of 0.3 seconds. We determine success by the robots ability to reach the target location. For each magnitude of force applied \mbox{([250N, 900N])}, we run 10 attempts and record the percentage of successfully completed tasks, as shown in Table \ref{tab:eval:pertunations} detailed in the next section.

\section{Results and Analysis}
\label{sec:results}

The section analyzes the experiments, discusses observations in relation to related works, and provides numerical evaluations for the kernel and the framework as a whole.

\subsection{Kernel Analysis}
\begin{figure}[!t]
    \centering
    \includegraphics[width=\linewidth]{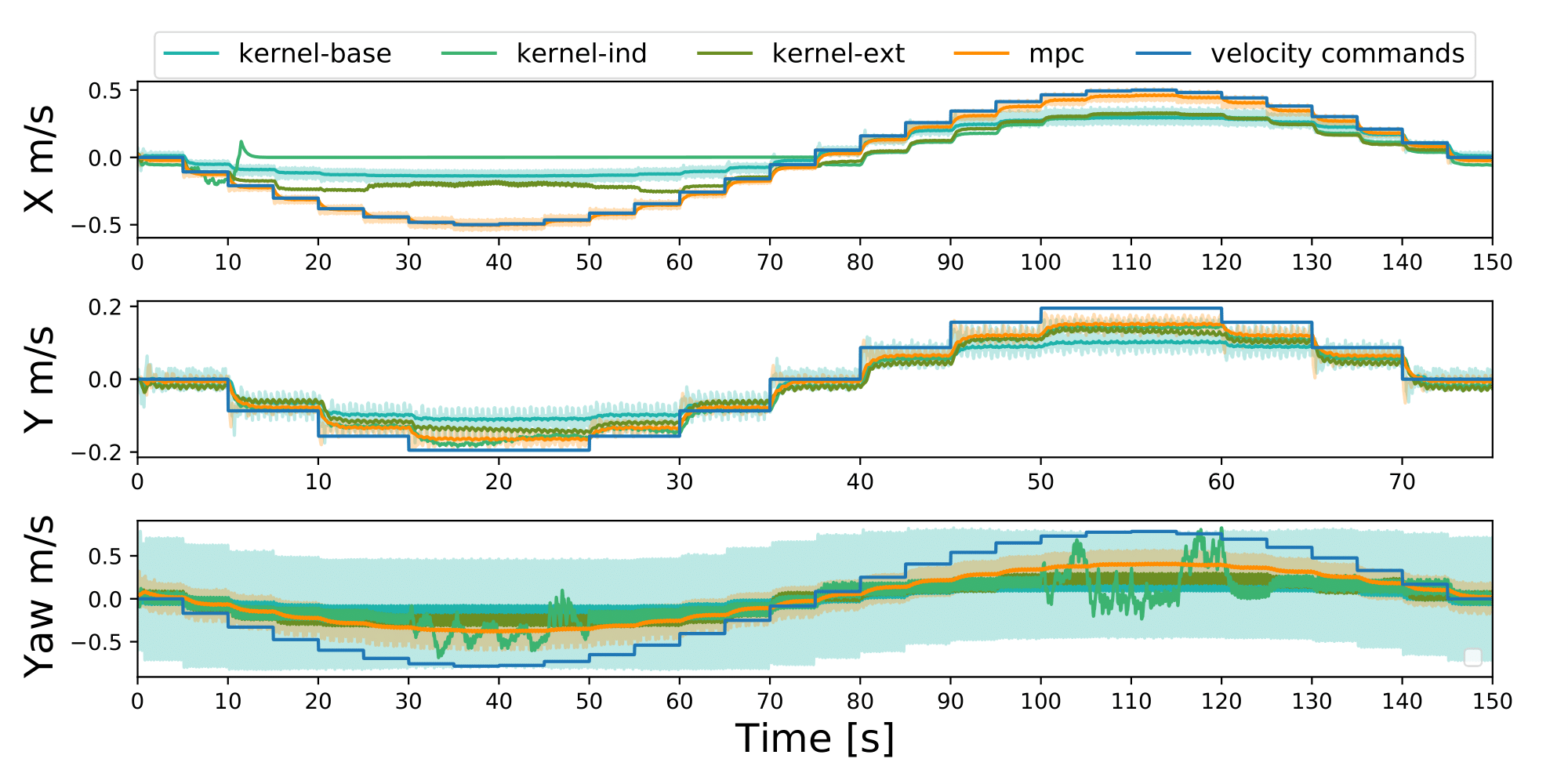}
    \vspace{-5mm}
    \caption{The realized velocities of the robot given velocity commands for the MPC controller and kernel variants.}
    \label{fig:velocity_tracking}
    \vspace{-5mm}
\end{figure}

To understand the degree kernel variants capture the characteristics and controllability of the MPC controller, we compare the velocity control performance exhibited on flat terrain, where a single velocity command is varied while the others are fixed to zero. 
Fig.~\ref{fig:velocity_tracking} demonstrates a performance gap between all variants and the MPC controller. The kernels cannot move at negative frontal velocities nor can they match the maximum lateral, angular and positive frontal velocities achieved with the MPC controller. In addition, the kernels experiences extremely high variance when turning, showing a significant performance gap in the realized yaw velocities. We observe no performance deterioration in \mbox{\textit{kernel-ext}} from \textit{kernel-base}, despite achieving lower validation loss (Table~\ref{tab:kernel_results_all_variants}). \textit{kernel-ind} is the weakest when moving at negative frontal velocities, but also experiences erratic behaviour when commanded with high yaw velocities.

\begin{figure}[!t]
    \centering
    \includegraphics[width=\linewidth]{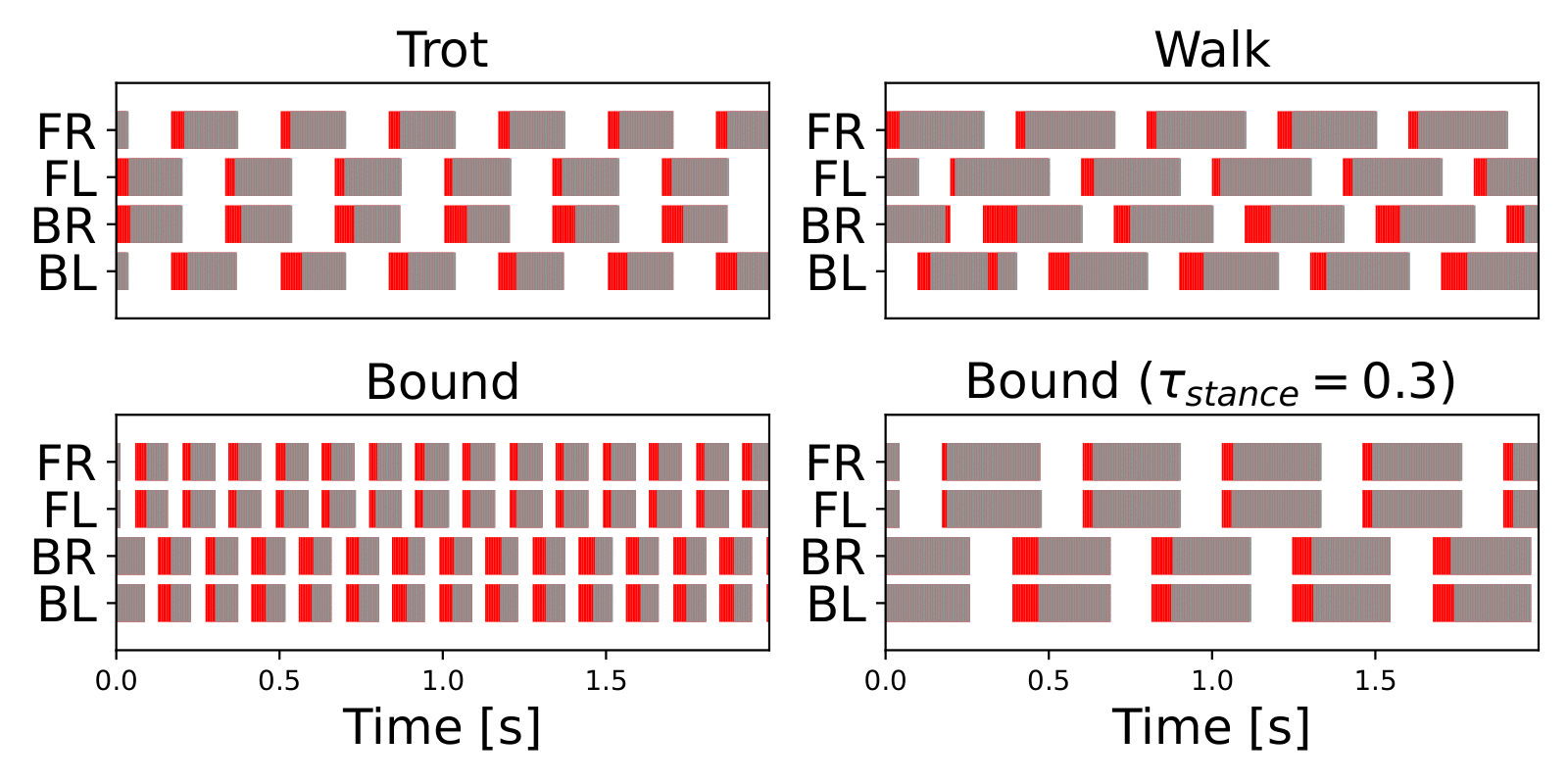}
    \vspace{-5mm}
    \caption{Zero shot gait patterns generated using Kernel-ind. Grey segments show the realized foot contacts, while the red segments show foot contact error against the contact schedule from the gait generator.  }
    \label{fig:gait_patterns_produced}
    \vspace{-5mm}
\end{figure}

\begin{figure}[!t]
    \centering
    \includegraphics[width=\linewidth, trim={0cm, 0cm, 0cm, 0.4cm}, clip=true]{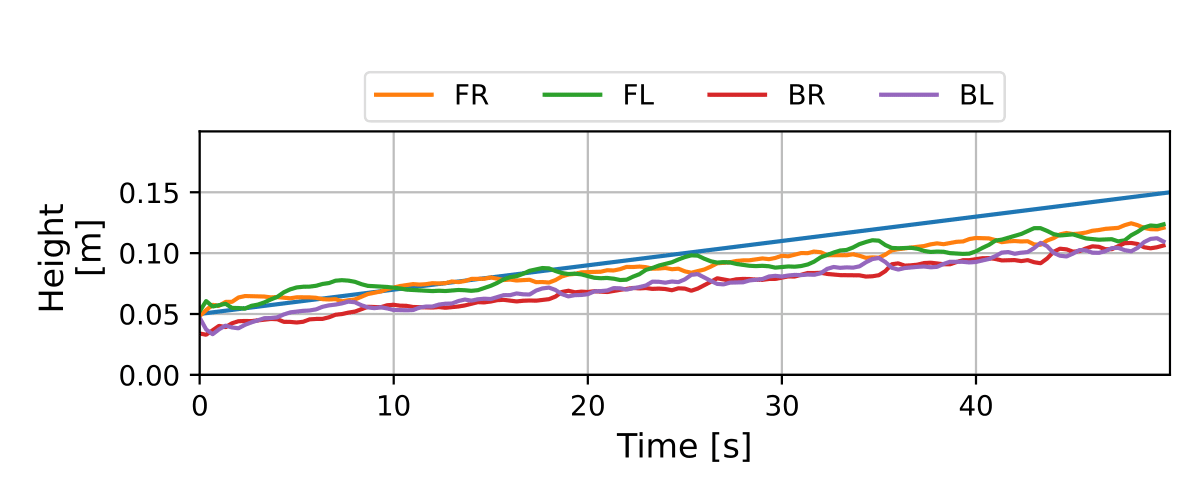}
    \vspace{-7mm}
    \caption{The peak realized height of each foot over a step cycle as the step height command increases (blue line), controlled using kernel-ext.  }
    \label{fig:step_heights_produced}
    \vspace{-5mm}
\end{figure}

\begin{figure}[!t]
    \centering
    \includegraphics[width=\linewidth, trim={0cm, 0cm, 0cm, 0.6cm}, clip=true]{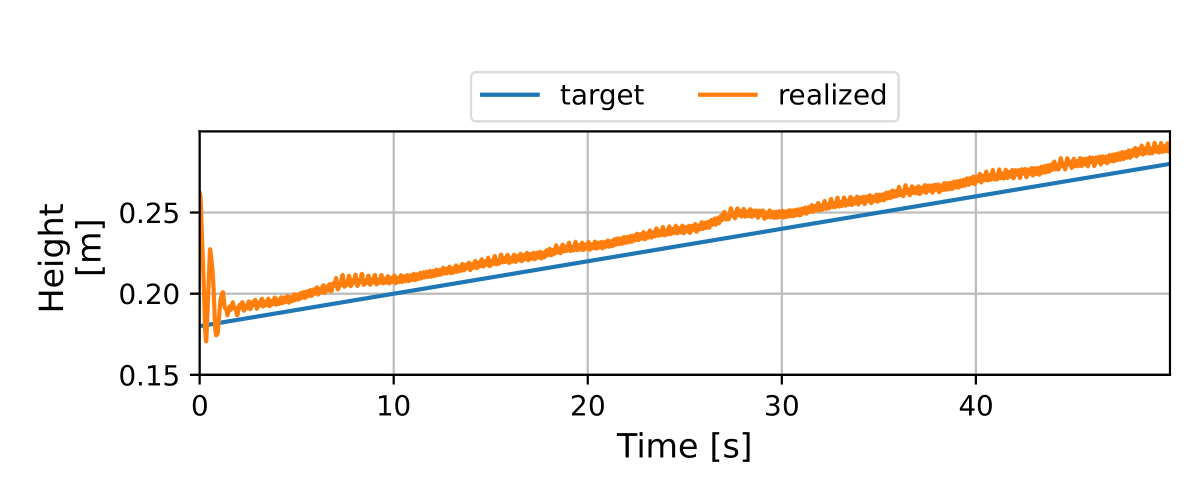}
    \vspace{-8mm}
    \caption{The realized height of the robots' base as the ride height command increases (blue line), controlled using kernel-ext.}
    \label{fig:ride_heights_produced}
    \vspace{-7mm}
\end{figure}

The variant, kernel-ind, demonstrates gait generalization capabilities, producing unseen gait patterns that result in effective locomotion. Fig.~\ref{fig:gait_patterns_produced} shows the production of walk and bound gaits, which were not provided during training. Although it produces these gaits, the kernel behaves undesirably when inputting high yaw commands as seen when executing the trot gait Fig.~\ref{fig:velocity_tracking}.

 Training of \textit{ kernel-ext} results a minimal increase in the validation loss (L1=$7.1e-4$, see Table \ref{tab:kernel_results_all_variants}), while allowing us to control the ride and step heights live, as shown in Fig.~\ref{fig:step_heights_produced} and Fig.~\ref{fig:ride_heights_produced} (The video can be found at \href{https://youtu.be/bUZJadWCRXU}{\small{\cblue{https://youtu.be/bUZJadWCRXU}}}).  We observed that the target step height and ride height commands are not realized precisely, although it clearly demonstrates the desired behaviour. Furthermore, we see greater inaccuracies in the realized steps heights, where the error increases as the target height increases.

\subsection{Kernel Results}

Our method demonstrates far superior results ($6.2e-4$ mean absolute error) compared to \cite{pretrained_rl}, which achieves a validation loss 0.007 (MSE), equivalent to 0.083 mean absolute error. Furthermore, our method yields a functional locomotion controller, as demonstrated by Fig.~\ref{fig:velocity_tracking}. The results (Table~\ref{tab:kernel_results_all_variants}) required training on 2.1 hours of locomotion data. Table \ref{tab:data_dep} shows the results of training with less data, determined by the number of target locations reached. The validation performance deteriorates as the number of targets decreases. However, training with only ten target locations (3.1 minutes), the kernel achieves a validation loss of $1e-3$, capable of producing locomotion simulation.

\begin{figure}[!t]
    \centering
    \includegraphics[width=\linewidth]{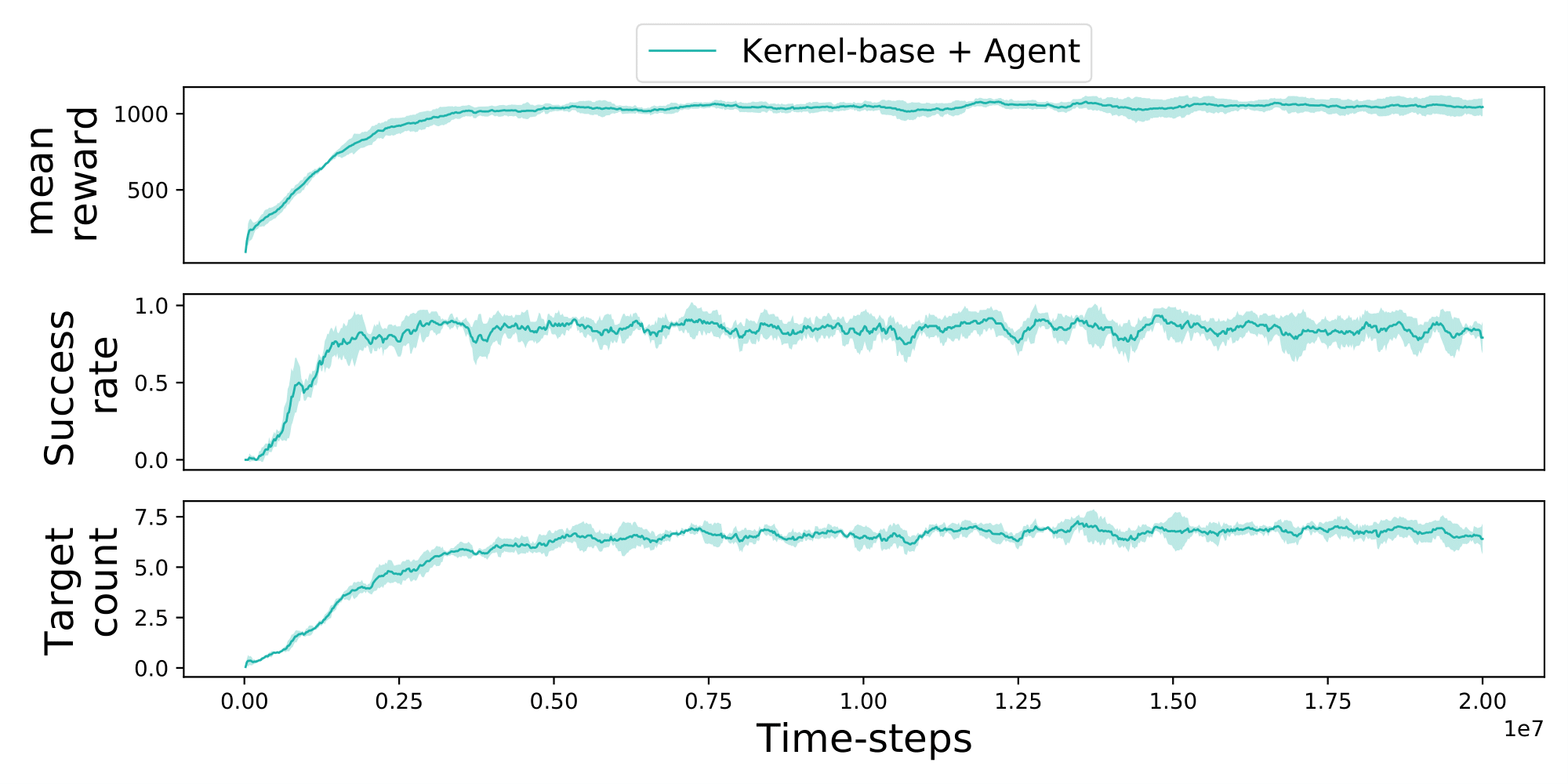}
    \caption{Training of the agent with \textit{kernel-base} to provide the priors, with the mean and standard deviation over four seeds. \label{fig:rl:rollout_traiing}}
    \vspace{-2mm}
\end{figure}

\begin{table}[!t]
\centering
\vspace{-2mm}
\caption{\label{tab:kernel_results_all_variants} Performance of kernel variants, showing the mean minimum validation loss and the standard deviation.}

\begin{tabular}{| c | c | c |  }
\hline
    \rowcolor{gray!30} Kernel-variant & Mean Validation Loss & Standard Deviation \\ \hline
    Kernel-base & $6.2e-4$ &  $6.9e-6$   \\\hline
    Kernel-ind & $7.2e-4$  &  $4.9e-6$  \\\hline
    Kernel-ext & $7.1e-4$  &  $1.0e-5$  \\\hline
\end{tabular}
\vspace{-0mm}
\end{table}

\begin{table}[!t]
\centering
\caption{\label{tab:data_dep} Kernel-base performance as the amount of data increases.}
\begin{tabular}{| c | c | c | c  |}
\hline
    \rowcolor{gray!30} Number of Targets & 10 & 25 & 50  \\ \hline
   Mean Validation Loss & $1e-3$ & $9.1e-4$ & $8.4e-4$   \\ \hline
   Standard Deviation & $3.1e-6$ & $9.8e-6$ & $5.3e-6$   \\ \hline

    \rowcolor{gray!30} Number of Targets &  100 & 200 & 400 \\ \hline
   Mean Validation Loss  & $7.7e-4$ & $6.7e-4$ & $6.2e-4$  \\ \hline
   Standard Deviation  & $6.2e-6$ & $5.5e-6$ & $6.9e-6$  \\ \hline
\end{tabular}
\vspace{-2mm}
\end{table}

\subsection{Residual Agent Analysis}

During training, we record the success rate, target count, and reward. An episode is considered successful after navigating to more than two target locations and not falling. The target count is the number of target locations reached with in a $60$s period.
The agent converges after only 7.5M timesteps (see Fig.~\ref{fig:rl:rollout_traiing}), showing significantly improved sample efficiency over other omnidirectional ResL methods: \cite{trajectory-adaption}, \cite{cpg-based}, and \cite{vae_humanoid}, requiring 250M, 100M, and 200M timesteps, respectively. This suggests deterministic reference motions, as provided by kernel-base and gait libraries, simplifies the learning scenario. Furthermore, our framework outperformed the kernel to seeded-agent framework \cite{pretrained_rl}, which 200M required timesteps.

\subsection{Residual Agent Results}

We measured the average reward per time-step, using the final reward function (Table~\ref{tab:rl:reward_function_final}); The success rate, defined as the proportion of complete runs (reaching all the targets), and the fall rate defined as the proportion of runs where the robot falls. Our framework (kernel+agent) demonstrates versatility outperforming the MPC controller used to train the kernel in every evaluation terrain with success rate of $93\%$ in the most challenging stairs terrain. The results are summarized in Table~\ref{tab:eva:sin}. Furthermore, it is more robust against perturbations, able to regularly recover its balance after perturbations of $800$N where the MPC controller fails.

\begin{table}[!t]
\centering
\begin{footnotesize}
\caption{Evaluation performance comparing locomotion controllers. \label{tab:eva:sin}}
\resizebox{0.95\linewidth}{!}{  
\begin{tabular}{|l|c|c|c|c|@{}}
\hline
    \rowcolor{gray!30}
    \textbf{Tabletop} & \textbf{Reward/steps} & \textbf{Num Targets}  & \textbf{Success Rate} \\ \hline

    MPC (0.2)  &  0.016$\pm$0.11   &  2.5$\pm$2.89  & \gradient{0.5}   \\ \hline
    Kernel   &  0.065$\pm$0.0085  & 2.25$\pm$1.5    & \gradient{0}     \\\hline
    \textbf{Kernel+Agent}  & 0.097$\pm$0.001     & 5$\pm$0.0    & \gradient{1}   \\ \hline
    
    \rowcolor{gray!30}
    \textbf{Seesaw} & \textbf{Reward/steps} & \textbf{Num Targets}  & \textbf{Success Rate}  \\ \hline
    MPC (0.2)  &  0.065$\pm$0.018   &  0.0$\pm$0.0   & \gradient{0}   \\\hline
    Kernel   &  0.043$\pm$0.00245   & 0.0$\pm$0.0    & \gradient{0}     \\\hline
    \textbf{Kernel+Agent}  &  0.091$\pm$0.0006   &  5$\pm$0.0    & \gradient{1}   \\\hline

    \rowcolor{gray!30}
    \textbf{Stairs} & \textbf{Reward/steps} & \textbf{Num Targets}  & \textbf{Success Rate}  \\ \hline
    MPC (0.2)  &  0.073$\pm$0.0022  &  0.0$\pm$0.0  & \gradient{0}   \\\hline
    Kernel   &  0.047$\pm$0.0019   &  0.0$\pm$0.0   & \gradient{0}   \\\hline
    \textbf{Kernel+Agent}   &  0.089$\pm$0.0024    &  4.75$\pm$1.0  & \gradient{0.9375}      \\ \hline

    \rowcolor{gray!30}
    \textbf{Sinusoidal} & \textbf{Reward/steps} & \textbf{Num Targets}  & \textbf{Success Rate}  \\ \hline
    MPC (0.2)  &  0.082$\pm$0.0057  &  3$\pm$1.83 & \gradient{0.25}    \\\hline
    Kernel   &  0.042$\pm$0.0021   &  0$\pm$0.0   &  \gradient{0}   \\\hline
    \textbf{Kernel+Agent}   &   0.089$\pm$0.0026  &   4.75$\pm$1.0  & \gradient{0.9375}   \\\hline
                          
\end{tabular}}
\end{footnotesize}
\vspace{-0mm}

\end{table}

\begin{table}[!t]
\centering
\begin{footnotesize}
\caption{\label{tab:eval:pertunations} Robustness against perturbations, using the MPC  with ($\tau_{stance}=$0.2).}
\resizebox{0.95\linewidth}{!}{  
\begin{tabular}{|c|c|c|c|c|c|c|c|c|c|}
    \hline
    \rowcolor{gray!30}Force (N) &250&300&350&400&450&500&550\\ \hline
    MPC & \gradient{1.0} & \gradient{0.8}  & \gradient{0.9}  & \gradient{0.7}  & \gradient{0.5}  & \gradient{0.3}  & \gradient{0.4} \\\hline
    Kernel & \gradient{1.0} & \gradient{1.0} & \gradient{1.0} & \gradient{0.8}  & \gradient{0.5}  & \gradient{0.3}  & \gradient{0.3} \\\hline
    Kernel+Agent& \gradient{1} & \gradient{1} & \gradient{1} & \gradient{1} & \gradient{0.9} & \gradient{0.8} & \gradient{0.9} \\
    \hline
    
    \rowcolor{gray!30}Force (N) &600&650&700&750&800&850&900\\ \hline
    MPC & \gradient{1.0} & \gradient{0.4}  & \gradient{0.1}  & \gradient{0.2}  & \gradient{0} & \gradient{0} & \gradient{0}\\\hline
    Kernel & \gradient{0} & \gradient{0} & \gradient{0} & \gradient{0} & \gradient{0} & \gradient{0} & \gradient{0}\\\hline
    Kernel+Agent & \gradient{0.8}  & \gradient{0.7}  & \gradient{0.4}  & \gradient{0.8}  & \gradient{0.6}  & \gradient{0.3}  & \gradient{0.2} \\
    \hline

\end{tabular}}
\end{footnotesize}
\vspace{-0mm}

\end{table}


\section{Conclusions}
\label{sec:conclusion}

In this work, we developed a ResL framework that is both sample efficient and highly controllable, providing omnidirectional locomotion at continuous velocities. We achieved this by providing deterministic trajectory priors using a NN trained on expert data collected from an MPC controller. Additionally, our residual agent applied positional trajectories without knowledge of the priors or the terrain. Through a set of  simulated scenarios, the framework demonstrated navigation on the most challenging terrains and demonstrated superior performance over the MPC controller used to train the kernel. Furthermore, the kernel exhibited gait generalization capabilities, producing locomotion for walk and bound gaits, when provided with only trot data.



For future work, we propose using the residual agent to adapt the trajectories produced for unseen gaits, to enable expert level control without any guidance directly from an expert controller. Additionally, we hypothesise the framework could exhibit greater robustness if the agent has direct control over the body height and the step height.

\section*{Acknowledgement}
This work is supported by  EU H2020 project Enhancing Healthcare with Assistive Robotic Mobile Manipulation (HARMONY, 101017008).









\bibliographystyle{IEEEtran}
\bibliography{main}

v

\end{document}